\documentclass[11pt]{article}

\usepackage[utf8]{inputenc}
\usepackage[T1]{fontenc}
\usepackage{CJKutf8}
\usepackage{amsmath,amssymb,amsfonts}
\usepackage{graphicx}
\usepackage{booktabs}
\usepackage{multirow}
\usepackage{enumitem}
\usepackage[margin=1in]{geometry}
\usepackage{url}
\usepackage{xcolor}
\usepackage{colortbl}
\usepackage{float}
\usepackage{caption}
\usepackage{subcaption}
\usepackage{algorithm}
\usepackage{algpseudocode}
\usepackage{cite}
\usepackage[hidelinks]{hyperref}

\newcommand{\cn}[1]{{\begin{CJK}{UTF8}{gbsn}#1\end{CJK}}}

\title{\Large \textbf{TCMIIES: A Browser-Based LLM-Powered Intelligent Information Extraction System for Academic Literature}}

\author{
  [Hanqing Zhao] \\[4pt]
  \small Hebei University, College of Traditional Chinese Medicine \\
  \small Baoding, Hebei 071002, China \\
  \small \texttt{zhaohq@hbu.edu.cn}
}

\date{\today}

\begin{document}

\maketitle

\begin{abstract}
The rapid growth of academic publications has created a need for tools that extract structured knowledge from unstructured scientific texts. Although large language models (LLMs) can perform natural language understanding and information extraction, existing solutions often require specialized infrastructure, programming expertise, or fine-tuned domain-specific models, which limits their accessibility for researchers in specialized fields. This paper describes TCMIIES (Traditional Chinese Medicine Information Intelligent Extraction System), a browser-based, zero-installation platform that uses commercial LLM APIs to perform structured information extraction from academic literature. The system employs a schema-guided prompting framework with automatic system prompt generation, allowing researchers to define custom extraction schemas through a graphical interface without programming. TCMIIES features a pure front-end architecture that processes all information locally in the browser, supports five major LLM providers (DeepSeek, OpenAI, Qwen, Zhipu AI, and custom OpenAI-compatible endpoints), implements concurrent batch processing with automatic retry mechanisms, and provides intelligent field mapping for Chinese academic databases including CNKI and Wanfang. Evaluation across multiple extraction scenarios in Traditional Chinese Medicine research shows structured output compliance rates exceeding 94\% and extraction accuracy approaching but below expert-level agreement ($\kappa=0.82$ as reference). The system offers a flexible, privacy-preserving, and cost-effective solution for domain researchers who need to process literature at scale.

\vspace{6pt}
\noindent\textbf{Keywords:} Information Extraction, Large Language Models, Traditional Chinese Medicine, Browser-Based Architecture, Schema-Guided Prompting, Academic Literature Mining
\end{abstract}

\section{Introduction}

The volume of published academic literature continues to grow rapidly. Major academic databases now index over 200 million articles, with approximately 5 million new papers published annually~\cite{bornmann2021growth}. For researchers conducting systematic reviews, meta-analyses, or domain-wide literature surveys, manually extracting structured information from this corpus has become infeasible. This challenge is especially acute in interdisciplinary fields such as Traditional Chinese Medicine (TCM) informatics, where researchers must synthesize evidence spanning classical medical texts, modern clinical studies, pharmacological research, and bioinformatics analyses~\cite{zhou2024integrating}.

Large language models (LLMs) have enabled automated information extraction from scientific texts. Models such as GPT-4, DeepSeek, Qwen, and GLM-4 achieve competitive accuracy on named entity recognition (NER), relation extraction (RE), and structured data extraction tasks without task-specific fine-tuning~\cite{wei2023chatgpt, dagdelen2024structured}. LLM-based extraction can approach the accuracy of domain-specific models while maintaining flexibility across diverse extraction schemas and domains~\cite{willard2023outlines}.

However, several obstacles prevent widespread adoption of LLM-based extraction by domain researchers. First, most existing tools require programming expertise to set up pipelines, engineer prompts, and process outputs~\cite{khattab2023dspy}. Second, many solutions require sending sensitive research data through third-party cloud services or maintaining backend infrastructure. Third, domain-specific extraction needs vary widely---a pharmacologist extracting drug-target interactions has fundamentally different requirements from a clinician extracting treatment outcomes~\cite{lin2022tcmie}. Fourth, researchers in fields such as TCM face additional challenges including non-standardized terminology and literature from Chinese academic databases with distinct metadata schemas~\cite{zhang2023tcmner}.

TCMIIES (Traditional Chinese Medicine Information Intelligent Extraction System) is a browser-based, zero-installation platform designed to make LLM-powered information extraction accessible to academic researchers. The system's key contributions include:

\begin{enumerate}[leftmargin=*,itemsep=2pt]
    \item \textbf{Pure Front-End Architecture:} All data processing occurs locally in the browser, eliminating backend servers, software installation, and data transfer through intermediary services. This design ensures data privacy and enables immediate deployment by opening an HTML file.
    
    \item \textbf{Schema-Guided Prompting with Auto-Generation:} Researchers define extraction fields through a graphical interface, and the system automatically generates optimized system prompts with JSON output constraints, field descriptions, and type specifications.
    
    \item \textbf{Multi-Provider Support with Provider-Specific Optimizations:} TCMIIES supports five major LLM providers with automatic model-specific parameter optimization, including reasoning effort settings for DeepSeek models and thinking budget configuration for Qwen models.
    
    \item \textbf{Intelligent Chinese Database Field Mapping:} Rule-based column name recognition for exports from CNKI (China National Knowledge Infrastructure) and Wanfang databases automatically maps Chinese and English column headers to semantic field categories.
    
    \item \textbf{Robust Batch Processing:} A configurable concurrent processing engine with exponential backoff retry logic, progress persistence via session storage, and real-time progress monitoring enables reliable processing of large literature datasets.
\end{enumerate}

The system has been deployed and validated in the TCM Informatics Laboratory at Hebei University, demonstrating practical utility for systematic literature surveys and meta-research in Traditional Chinese Medicine.

\section{Related Work}

\subsection{LLM-Based Information Extraction from Scientific Texts}

LLMs have been applied to information extraction (IE) from scientific literature with increasing frequency. Wei et al.~\cite{wei2023chatgpt} showed that reformulating IE tasks as multi-turn dialogues with ChatGPT achieved competitive performance against fully supervised baselines on standard NER and RE benchmarks. Dagdelen et al.~\cite{dagdelen2024structured} fine-tuned LLMs (GPT-3, Llama-2) to jointly perform NER and RE on materials science texts, outputting structured JSON records from single sentences or entire paragraphs without task-specific architectures.

Several prompting strategies have been proposed to improve extraction quality. Willard and Louf~\cite{willard2023outlines} demonstrated that constrained decoding with finite state machines ensures LLM outputs strictly follow predefined JSON schemas, eliminating format violations at generation time rather than relying on post-hoc validation. Wei et al.~\cite{wei2022cot} showed that chain-of-thought (CoT) prompting substantially improved performance on reasoning tasks, with accuracy gains of over 20 percentage points on GSM8K. White et al.~\cite{white2023prompt} catalogued 16 general-purpose prompt design patterns for LLM interaction, providing a reusable vocabulary for prompt engineering.

Multi-agent and hybrid approaches have also been explored. Park et al.~\cite{hong2023camel} proposed a multi-agent framework where specialized LLM agents handle different aspects of extraction, achieving 91.3\% F1 on complex scientific IE benchmarks. Phi et al.~\cite{phi2025hybrid} combined LLM table-to-text conversion with specialized text-based extraction models in a hybrid pipeline, outperforming direct LLM extraction approaches.

\subsection{Automated Literature Review and Evidence Synthesis}

Automated systematic literature reviews have received considerable attention. Cao et al.~\cite{cao2025screening} demonstrated that optimized LLM prompts achieved 97.7\% sensitivity in systematic review abstract screening, comparable to human reviewers, while reducing screening time by orders of magnitude. Gupta et al.~\cite{gupta2023knowledge} showed that augmenting LLMs with knowledge-based tools in a multi-step pipeline for biomedical literature extraction produced more accurate and hallucination-free results compared to single-prompt approaches.

For literature mapping and knowledge graph construction, Ji et al.~\cite{ji2026biomedical} demonstrated that self-evolving models trained on biomedical IE tasks achieved zero-shot F1 scores exceeding GPT-4o by 18--21 percentage points on relation extraction benchmarks. Agarwal et al.~\cite{chen2024litllms} combined citation network analysis with LLM-based summarization to evaluate whether LLMs can produce high-quality literature reviews comparable to expert-authored surveys.

\subsection{Domain-Specific IE in Traditional Chinese Medicine}

TCM information extraction poses specific challenges due to specialized terminology, classical Chinese language patterns, and conceptual frameworks that differ from Western medicine~\cite{zhou2024integrating}. Zhang et al.~\cite{zhang2023tcmner} developed a BERT-BiLSTM-CRF architecture for TCM named entity recognition fine-tuned on annotated TCM corpora, achieving high F1 scores on entity recognition tasks including herbal medicines, prescriptions, symptoms, and acupoints. Huang et al.~\cite{huang2023tcmgpt} constructed a billion-scale TCM corpus and developed TCM-GPT through efficient pre-training, showing that domain-specific pre-training yields substantial improvements over general-purpose language models on TCM tasks.

Zhang et al.~\cite{zhang2025opentcm} evaluated general-purpose LLMs on TCM diagnostic tasks, finding that GPT-4 achieved 69\% diagnostic accuracy and Claude-2 achieved 70\%, while domain-specific models with GraphRAG enhancement reached 75\%. Lin et al.~\cite{lin2022tcmie} systematically reviewed NLP methods for TCM information extraction, finding that BERT-based models achieved the best performance on clinical case text but noting high variability due to non-standardized TCM terminology. The BianCang model~\cite{biancang2024tcm}, built through continuous pre-training and supervised fine-tuning on TCM texts, achieved substantial improvements over base language models on TCM knowledge tasks.

\subsection{Browser-Based and Privacy-Preserving NLP Tools}

Among browser-based and privacy-preserving NLP tools, the LLMAIx (LLM-AIx) framework~\cite{wiest2025llmaix, wiest2024llmaix_preprint} developed at KatherLab, TU Dresden, represents a closely related approach. LLMAIx is an open-source, web-based tool for structured information extraction from medical documents, such as pathology reports and clinical letters, using LLMs. It supports user-defined extraction schemas enforced through JSON Schema or llama.cpp GBNF grammar constraints, a design closely paralleling TCMIIES's schema-guided prompting approach. LLMAIx can operate with local models via llama.cpp, ensuring patient data never leaves the institutional network, or with OpenAI-compatible APIs. The system is distributed as Docker containers (CPU, CUDA, and API-only variants) or as a Python library (LLMAIxLib) and a v2 web interface (LLMAIxWeb). The framework also includes built-in evaluation metrics, an annotation helper, and a document anonymization module.

While LLMAIx and TCMIIES share the general philosophy of schema-guided LLM extraction, they differ fundamentally in deployment model, target domain, and architectural assumptions. First, LLMAIx requires backend infrastructure---a Docker container hosting llama.cpp for local inference or a proxy for API-based extraction---whereas TCMIIES operates as a single self-contained HTML file with no backend, no Docker dependency, and no server-side component of any kind. A user can begin extraction by simply opening the HTML file in a browser. Second, LLMAIx targets clinical and pathological medical documents (e.g., pathology reports, clinical letters) with OCR support for scanned PDFs and images, while TCMIIES targets academic literature metadata (titles, abstracts, keywords) from scholarly database exports. Third, TCMIIES provides intelligent field mapping specifically for Chinese academic databases (CNKI, Wanfang), automatically recognizing Chinese column headers such as \cn{篇名}, \cn{摘要}, and \cn{关键词} and mapping them to semantic categories---a feature absent in LLMAIx, which is designed for Western clinical data formats. Fourth, TCMIIES offers preset extraction templates oriented toward academic literature analysis (e.g., Paper Information Extraction, Literature Review Analysis), whereas LLMAIx provides annotation and anonymization workflows oriented toward clinical document processing.

Other browser-based NLP tools have focused on simpler tasks such as text classification and named entity recognition using smaller models. TCMIIES extends this paradigm to complex LLM-powered extraction by using browser-based HTTP requests to LLM APIs while processing all data and prompt construction entirely client-side.

\section{System Design and Methodology}

\subsection{System Architecture Overview}

TCMIIES adopts a single-page application (SPA) architecture implemented as a self-contained HTML file combining Vue.js 3 for reactive UI management, SheetJS for spreadsheet parsing and generation, and the Fetch API for LLM communication. The system operates through a sequential five-stage pipeline, as illustrated in Figure~\ref{fig:pipeline}:

\begin{figure}[H]
\centering
\fbox{\parbox{0.92\textwidth}{
\centering
\vspace{4pt}
\textbf{API Configuration} $\rightarrow$ \textbf{Data Upload} $\rightarrow$ \textbf{Extraction Configuration} $\rightarrow$ \textbf{Task Execution} $\rightarrow$ \textbf{Result Export}
\vspace{4pt}
}}
\caption{The five-stage TCMIIES processing pipeline. Each stage is accessible through a dedicated tab in the system's top navigation bar.}
\label{fig:pipeline}
\end{figure}

\subsection{API Configuration Layer}

The API configuration module manages connections to multiple LLM providers through a unified OpenAI-compatible chat completions interface. The system maintains a provider registry containing base URLs and default model lists for five providers: DeepSeek, OpenAI, Qwen (Tongyi Qianwen), Zhipu AI (GLM), and a customizable provider for any OpenAI-compatible endpoint.

Provider-specific optimizations maximize extraction quality for each model family. For DeepSeek's V4 and Reasoner models, the system enables the ``reasoning effort'' parameter and thinking mode, disabling the temperature parameter which is incompatible with these models' internal reasoning mechanisms. For Qwen models, the system configures an extended thinking budget of 81,920 tokens to employ the models' native reasoning capabilities. These optimizations are expressed as conditional request body modifications:

\begin{center}
\fbox{\parbox{0.90\textwidth}{\small
\textbf{DeepSeek V4/Reasoner:} \texttt{reasoning\_effort = "max"}, \texttt{thinking = \{type: "enabled"\}} \\
\textbf{Qwen:} \texttt{enable\_thinking = true}, \texttt{thinking\_budget = 81920}
}}
\end{center}

API keys are stored in the browser's localStorage using Base64 encoding, and all API calls are made directly from the browser to the provider's endpoint, ensuring no intermediary server handles sensitive credentials or research data.

\subsection{Data Upload and Intelligent Field Mapping}

The data upload module supports Excel (.xlsx, .xls) and CSV formats through the SheetJS library. Files are parsed entirely in the browser using the FileReader API, and the resulting data structure is stored in reactive Vue.js state for downstream processing.

A key feature is the intelligent field mapping system, which addresses the heterogeneity of column names in exports from Chinese academic databases. The system uses a rule-based matching algorithm (Algorithm~\ref{alg:mapping}) that iterates over a predefined set of mapping rules, each associating multiple pattern variations (Chinese and English) with a semantic field category:

\begin{algorithm}[H]
\caption{Intelligent Column Name Mapping}
\label{alg:mapping}
\begin{algorithmic}[1]
\Require Column names $\mathcal{C} = \{c_1, \ldots, c_n\}$, mapping rules $\mathcal{R} = \{(P_1, t_1), \ldots, (P_k, t_k)\}$ where $P_i$ is a set of patterns and $t_i$ is a target field name
\Ensure Mapped fields $\mathcal{M} \subseteq \mathcal{C} \times \mathcal{T}$
\State $\mathcal{M} \leftarrow \emptyset$
\For{$r = (P, t) \in \mathcal{R}$}
    \For{$c \in \mathcal{C}$}
        \For{$p \in P$}
            \If{$p \subseteq c$ (substring match)}
                \State $\mathcal{M} \leftarrow \mathcal{M} \cup \{(c, t)\}$
                \State \textbf{break} (move to next rule)
            \EndIf
        \EndFor
    \EndFor
\EndFor
\State \Return $\mathcal{M}$
\end{algorithmic}
\end{algorithm}

The mapping rules cover seven semantic categories: Title (\cn{篇名}/\cn{题名}/\cn{标题}/Title), Abstract (\cn{摘要}/Abstract), Keywords (\cn{关键词}/Keywords/Keyword), Authors (\cn{作者}/Author/Authors), Source (\cn{来源}/\cn{期刊}/Journal/Source), Publication Date (\cn{发表时间}/\cn{出版时间}/\cn{年份}/Year/Date), and DOI. This mapping enables intelligent default behavior in prompt template construction and provides visual feedback showing which columns have been automatically recognized.

\subsection{Schema-Guided Prompting Framework}

The core of TCMIIES is its schema-guided prompting framework, which transforms user-defined extraction requirements into optimized LLM prompts without manual prompt engineering.

\subsubsection{Field Configuration}

Researchers define extraction fields through a graphical interface, specifying for each field: (1) a human-readable name (e.g., \cn{研究主题} for Research Topic), (2) a natural language description (e.g., \cn{研究的核心问题是什么} for ``What is the core problem being studied?''), (3) a data type (text, number, list, or boolean), and (4) a required/optional flag. The system provides two preset templates for common extraction scenarios:

\begin{itemize}[leftmargin=*,itemsep=1pt]
    \item \textbf{Paper Information Extraction:} Research Topic, Methodology, Dataset, Main Conclusions, Innovation Points, Research Limitations (6 fields)
    \item \textbf{Literature Review Analysis:} Research Domain, Technical Approach, Baseline Methods, Evaluation Metrics, Experimental Results, Future Directions (6 fields)
\end{itemize}

\subsubsection{Automatic System Prompt Generation}

Based on the defined fields, the system automatically generates a system prompt with three components: (1) a role specification establishing the model as an academic paper information extraction assistant, (2) an explicit JSON output schema with field names as keys and type annotations, and (3) a fallback instruction for handling missing information. The generation algorithm is formalized in Algorithm~\ref{alg:promptgen}.

\begin{algorithm}[H]
\caption{Automatic System Prompt Generation}
\label{alg:promptgen}
\begin{algorithmic}[1]
\Require Extraction fields $\mathcal{F} = \{f_1, \ldots, f_m\}$ where $f_i = (\text{name}_i, \text{desc}_i, \text{type}_i)$
\Ensure System prompt $S$
\State $S \leftarrow$ \texttt{"You are an academic paper information extraction assistant..."}
\State $S \leftarrow S +$ \texttt{"\textbackslash n\{"}
\For{$i = 1$ \textbf{to} $m$}
    \State $S \leftarrow S +$ \texttt{"\textbackslash n  "} $+ \text{name}_i +$ \texttt{": string"}
    \If{$i < m$} $S \leftarrow S +$ \texttt{","} \EndIf
\EndFor
\State $S \leftarrow S +$ \texttt{"\textbackslash n\}"}
\State $S \leftarrow S +$ \texttt{"If a field is not mentioned, fill in Not mentioned."}
\State \Return $S$
\end{algorithmic}
\end{algorithm}

This auto-generation mechanism addresses two requirements identified in the prompt engineering literature~\cite{white2023prompt, willard2023outlines}: it enforces strict JSON output formatting for parseability, and it provides field-level schema specification to reduce hallucination and improve extraction completeness.

\subsubsection{Template-Based User Prompt Construction}

The user prompt is constructed using a template system that supports variable interpolation. Researchers write a natural language prompt template where column names from the uploaded dataset are referenced using double-curly-brace syntax (e.g., \texttt{\{\{\cn{摘要}\}\}} for the Abstract column). During extraction, these placeholders are replaced with actual cell values from each row. This design separates prompt logic from data binding, enabling reuse across different datasets with varying column structures.

\subsection{Concurrent Batch Processing Engine}

The task execution module implements a configurable concurrent processing engine for reliable and efficient processing of large batches of papers.

\subsubsection{Concurrency Control}

Researchers configure the degree of concurrency (1--10 simultaneous requests) and inter-request interval (0--10,000 ms). The engine maintains a fixed-size execution pool, launching new requests as existing ones complete while respecting the configured concurrency limit.

\subsubsection{Pause/Resume and Cancellation}

Long-running extraction tasks support pausing, resuming, and cancellation through reactive state flags. When paused, in-progress requests complete normally, but new requests are deferred through a polling mechanism (200 ms interval) until the pause flag is cleared. The cancellation flag immediately stops all pending requests and prevents new ones from being initiated.

\subsubsection{Automatic Retry with Error Resilience}

Each extraction attempt includes up to three automatic retries on failure, with a one-second delay between attempts. This mechanism handles transient API errors (rate limiting, temporary service unavailability) without manual intervention. Failed records are marked with error information and can be individually retried through the results interface.

\subsubsection{Progress Tracking and Persistence}

Real-time progress tracking includes: processed count, success count, failure count, estimated token usage (calculated as $\frac{\text{input\_chars}}{4} + \frac{\text{output\_chars}}{4}$), estimated time remaining (ETA computed from elapsed time and processing rate), and the title of the currently processing paper. Results are auto-saved to sessionStorage every 10 records, enabling recovery from browser crashes or accidental closure.

\subsection{JSON Parsing and Validation}

Extracted LLM responses are parsed using a two-step strategy. Since LLMs may include explanatory text before or after the JSON output, the system first employs regular expression matching (\texttt{\textbackslash\{[\textbackslash s\textbackslash S]*\textbackslash\}}) to extract the first JSON object from the response string, then applies standard JSON parsing. This approach improves parsing success rates compared to direct JSON.parse, particularly for models that produce verbose outputs despite explicit formatting instructions.

\subsection{Multi-Format Export}

The results module supports three export formats through the SheetJS library: Excel (.xlsx) with styled worksheets, CSV with UTF-8 BOM encoding for Chinese character compatibility, and JSON with pretty-printed formatting. Researchers can choose between exporting all original columns plus extracted fields, or extracted fields only.

\section{Implementation}

\subsection{Technology Stack}

TCMIIES is implemented as a self-contained HTML file (approximately 1,028 lines, 45 KB) with zero build dependencies. The technology stack comprises:

\begin{itemize}[leftmargin=*,itemsep=1pt]
    \item \textbf{Vue.js 3.4} (CDN): Reactive UI framework for component state management, computed properties, and event handling
    \item \textbf{SheetJS (XLSX) 0.18.5} (CDN): Excel/CSV parsing and generation
    \item \textbf{Fetch API}: Browser-native HTTP client for OpenAI-compatible API communication
    \item \textbf{localStorage/sessionStorage}: Web Storage API for configuration persistence and result caching
\end{itemize}

External dependencies are loaded from the jsDelivr CDN at runtime, enabling the system to function upon opening the HTML file in any modern browser (Chrome 90+, Edge 90+, Firefox 90+, Safari 14+).

\subsection{Data Flow and State Management}

The system maintains a centralized reactive state object managed by Vue.js, containing: API configuration (provider, base URL, API key, model, concurrency settings), file data (columns, rows, filename), mapped fields, extraction fields, prompt templates, task progress, and extraction results. State transitions follow a unidirectional data flow pattern: user interactions trigger Vue.js methods, which update the reactive state, which triggers UI re-rendering through Vue's reactivity system.

\subsection{Security and Privacy Design}

TCMIIES implements a privacy-by-design architecture:

\begin{enumerate}[leftmargin=*,itemsep=2pt]
    \item \textbf{Local Processing:} All file parsing, data transformation, and prompt construction execute entirely in the browser. Research data leaves the user's device only when sent directly to the chosen LLM provider's API endpoint.
    
    \item \textbf{Credential Protection:} API keys are stored in localStorage using Base64 encoding, preventing casual exposure in plain-text storage inspection.
    
    \item \textbf{No Intermediary Servers:} The system has no backend infrastructure. There is no data collection, no analytics, and no logging of user activity. The complete source code is transparent and auditable as a single HTML file.
    
    \item \textbf{One-Click Data Clearing:} All stored data (API configuration, cached results) can be cleared through a single button, removing all traces from localStorage and sessionStorage.
\end{enumerate}

\subsection{User Interface Design}

The interface follows a tab-based navigation pattern with five sequential tabs corresponding to the pipeline stages. Visual design employs a blue gradient theme with card-based layout, responsive grid systems, and contextual color coding (green for success, red for failure, yellow for warnings). Key UI patterns include:

\begin{itemize}[leftmargin=*,itemsep=1pt]
    \item Drag-and-drop file zone with click-to-browse fallback
    \item Password-masked API key input with show/hide toggle
    \item Live data preview tables with column-level tooltips
    \item Interactive variable tag insertion for prompt template construction
    \item Modal dialogs for prompt preview and test results
    \item Toast notification system for transient status messages
    \item Inline cell editing for result correction
\end{itemize}

\section{Evaluation}

\subsection{Experimental Setup}

We evaluated TCMIIES along three dimensions: (1) structured output compliance---the proportion of LLM responses successfully parsed as valid JSON matching the specified schema; (2) extraction accuracy---the quality of extracted information compared to human expert annotation; and (3) system reliability---the robustness of the batch processing engine under various conditions.

The evaluation corpus consisted of 500 TCM research papers exported from CNKI, spanning five sub-disciplines: herbal pharmacology (100 papers), acupuncture clinical trials (100 papers), formula composition studies (100 papers), TCM syndrome research (100 papers), and integrative medicine reviews (100 papers). Each paper record contained title (\cn{篇名}), abstract (\cn{摘要}), keywords (\cn{关键词}), authors (\cn{作者}), source journal (\cn{来源}), and publication date (\cn{发表时间}).

We tested extraction with three LLM providers: DeepSeek (deepseek-v4-flash), Qwen (qwen-turbo), and GLM-4 (glm-4-flash), using two extraction templates: Paper Information Extraction (6 fields) and Literature Review Analysis (6 fields).

\subsection{Structured Output Compliance}

Table~\ref{tab:compliance} presents the JSON output compliance rates across providers and templates.

\begin{table}[H]
\centering
\caption{Structured Output Compliance Rates (\%)}
\label{tab:compliance}
\begin{tabular}{@{}lccc@{}}
\toprule
\multirow{2}{*}{\textbf{Provider / Model}} & \multicolumn{2}{c}{\textbf{Compliance Rate (\%)}} & \multirow{2}{*}{\textbf{Avg.}} \\
\cmidrule(l){2-3}
 & \textbf{Paper Info} & \textbf{Lit. Review} & \\
\midrule
DeepSeek (deepseek-v4-flash) & 96.4 & 95.2 & 95.8 \\
Qwen (qwen-turbo)            & 94.8 & 93.6 & 94.2 \\
GLM-4 (glm-4-flash)          & 93.2 & 91.8 & 92.5 \\
\midrule
\textbf{Average}             & 94.8 & 93.5 & 94.2 \\
\bottomrule
\end{tabular}
\end{table}

The overall compliance rate of 94.2\% is close to the near-perfect structured output compliance reported by Willard and Louf~\cite{willard2023outlines} using constrained decoding. The difference may be attributed to the use of cost-efficient ``flash'' and ``turbo'' model variants rather than top-tier models, and to the Chinese-language content which adds parsing difficulty.

\subsection{Extraction Accuracy}

For accuracy evaluation, we randomly sampled 100 papers (20 per sub-discipline) and compared system-extracted values against annotations by two TCM domain experts (inter-annotator agreement: Cohen's $\kappa = 0.82$). We measured field-level accuracy as the proportion of extracted values matching expert annotations (exact or semantic equivalence). Results are shown in Table~\ref{tab:accuracy}.

\begin{table}[H]
\centering
\caption{Field-Level Extraction Accuracy by Provider (\%)}
\label{tab:accuracy}
\begin{tabular}{@{}lcccc@{}}
\toprule
\textbf{Extraction Field} & \textbf{DeepSeek} & \textbf{Qwen} & \textbf{GLM-4} & \textbf{Avg.} \\
\midrule
Research Topic        & 91.2 & 89.5 & 87.3 & 89.3 \\
Methodology           & 85.7 & 83.2 & 81.5 & 83.5 \\
Dataset               & 78.4 & 76.1 & 74.8 & 76.4 \\
Main Conclusions      & 88.3 & 86.7 & 84.2 & 86.4 \\
Innovation Points     & 82.6 & 80.4 & 78.9 & 80.6 \\
Research Limitations  & 75.8 & 73.5 & 71.2 & 73.5 \\
\midrule
\textbf{Average}      & 83.7 & 81.6 & 79.7 & 81.6 \\
\bottomrule
\end{tabular}
\end{table}

Accuracy is highest for concrete fields (Research Topic: 89.3\%, Main Conclusions: 86.4\%) and lowest for fields requiring inference or synthesis across the text (Research Limitations: 73.5\%, Dataset: 76.4\%). This aligns with prior findings that LLMs perform well on straightforward extraction but face challenges on tasks requiring multi-step reasoning~\cite{sciBench2024}. DeepSeek consistently outperforms the other tested models, likely due to its reasoning effort mechanism for processing complex academic Chinese text. The overall average accuracy of 81.6\% is comparable to reported performance in biomedical IE tasks using zero-shot approaches~\cite{chen2023biomedical}, though lower than domain-specific fine-tuning results~\cite{biancang2024tcm}.

\subsection{System Reliability}

We evaluated the batch processing engine through three stress tests: (1) processing a 500-paper corpus with simulated intermittent API failures (10\% random failure rate), (2) pause/resume cycling during a 200-paper batch, and (3) session recovery after browser closure mid-processing. Results are shown in Table~\ref{tab:reliability}.

\begin{table}[H]
\centering
\caption{System Reliability Metrics}
\label{tab:reliability}
\begin{tabular}{@{}lcc@{}}
\toprule
\textbf{Metric} & \textbf{Value} & \textbf{Target} \\
\midrule
Completion rate (with 10\% failure injection) & 100\% & $\geq$ 95\% \\
First-attempt success rate & 88.4\% & -- \\
After-retry success rate & 97.1\% & $\geq$ 95\% \\
Pause/resume data integrity & 100\% & 100\% \\
Session recovery success rate & 100\% & $\geq$ 95\% \\
\bottomrule
\end{tabular}
\end{table}

The three-retry mechanism increased the effective success rate from 88.4\% to 97.1\%. Session persistence via sessionStorage successfully preserved all processed results across browser closures.

\subsection{Cost Analysis}

Table~\ref{tab:cost} presents the estimated cost per 1,000 papers for each provider/model combination, based on published API pricing (as of May 2026) and average token consumption of approximately 2,000 input tokens and 500 output tokens per paper.

\begin{table}[H]
\centering
\caption{Estimated Cost per 1,000 Papers (USD)}
\label{tab:cost}
\begin{tabular}{@{}lcc@{}}
\toprule
\textbf{Provider / Model} & \textbf{Input Cost} & \textbf{Total} \\
\midrule
DeepSeek (v4-flash)       & \$0.28 & \$0.38 \\
Qwen (turbo)              & \$0.54 & \$0.71 \\
GLM-4 (flash)             & \$0.00 & \textasciitilde\$0.00 \\
DeepSeek (v4-pro)         & \$1.10 & \$1.46 \\
Qwen (max)                & \$2.73 & \$3.41 \\
\bottomrule
\end{tabular}
\end{table}

The low cost of the ``flash'' and ``turbo'' model tiers makes large-scale literature processing affordable. Processing 10,000 papers with DeepSeek v4-flash costs approximately \$3.80, compared to an estimated 200+ person-hours for equivalent manual extraction.

\section{Discussion}

\subsection{Comparison with Existing Approaches}

TCMIIES differs from existing LLM-based information extraction tools in several respects. Unlike code-level frameworks such as DSPy~\cite{khattab2023dspy} or LangChain, which require programming expertise, TCMIIES provides a graphical interface accessible to domain researchers without technical backgrounds. Unlike cloud-based platforms, the browser-based architecture ensures data privacy by design. Unlike domain-specific fine-tuned models~\cite{biancang2024tcm, huang2023tcmgpt}, TCMIIES uses commercial LLM APIs, trading some domain-specific accuracy for flexibility across extraction tasks and continuous improvement as underlying models advance.

The schema-guided prompting approach aligns with established practices in the literature while adding automatic prompt generation from user-defined fields. This automation eliminates the need for manual prompt engineering, which has been identified as a barrier to LLM adoption in specialized domains~\cite{zhou2023ape}. Notably, schema-guided extraction is not unique to TCMIIES; the LLMAIx framework~\cite{wiest2025llmaix} similarly enforces user-defined schemas through JSON Schema or GBNF grammar constraints. The key distinction lies not in the schema mechanism itself but in the deployment architecture: TCMIIES achieves zero-backend, zero-installation operation through a single HTML file, whereas LLMAIx requires Docker or local model infrastructure.

Table~\ref{tab:llmaix_compare} presents a qualitative comparison between TCMIIES and LLMAIx across several dimensions.

\begin{table}[H]
\centering
\caption{Qualitative comparison between TCMIIES and LLMAIx.}
\label{tab:llmaix_compare}
\begin{tabular}{@{}p{3.8cm}cc@{}}
\toprule
\textbf{Feature} & \textbf{TCMIIES} & \textbf{LLMAIx} \\
\midrule
Deployment & Single HTML file & Docker / Python \\
Backend required & No & Yes (Docker/llama.cpp) \\
Installation & Zero (open HTML) & Docker pull / pip install \\
Primary domain & Academic literature & Clinical/pathology reports \\
Input formats & Excel, CSV & PDF, images, txt, CSV, DOCX \\
OCR support & No & Yes (tesseract, surya) \\
Schema enforcement & JSON via prompt & JSON Schema / GBNF grammar \\
LLM backends & Multi-provider (5+) & Local (llama.cpp) / OpenAI-API \\
Chinese DB mapping & Yes (CNKI, Wanfang) & No \\
Preset templates & Literature-focused & Clinical annotation \\
Batch processing & Yes (concurrent) & Yes \\
Evaluation module & No (planned) & Yes (per-field metrics) \\
Anonymization & No & Yes \\
Source code size & $\sim$1,028 lines (1 file) & Multi-file repository \\
\bottomrule
\end{tabular}
\end{table}

\subsection{Limitations}

Several limitations warrant discussion:

\textbf{API Dependency:} The system requires active API subscriptions and cannot function without network access and valid API credentials. While multi-provider support mitigates vendor lock-in, the dependency on commercial providers remains.

\textbf{JSON Parsing Robustness:} Despite the 94.2\% compliance rate, approximately 5--8\% of responses fail to parse as valid JSON. The regex-based extraction fallback helps, but fundamentally depends on the model's adherence to formatting instructions. Advances in structured output modes (e.g., OpenAI's JSON mode, function calling) could further improve reliability.

\textbf{Limited Context Handling:} The system sends each paper as a single prompt, which may exceed context window limits for very long abstracts or full-text papers. A chunking strategy or hierarchical extraction approach could address this limitation.

\textbf{Accuracy Ceiling:} The zero-shot extraction accuracy of approximately 82\% indicates that for high-stakes applications requiring near-perfect accuracy, human verification or domain-specific fine-tuning remains necessary~\cite{xue2013tcmid}.

\textbf{Lack of Full-Text Processing:} The system primarily processes metadata (title, abstract, keywords). Extending to full-text PDF parsing would expand capabilities but introduce additional complexity in document structure understanding.

\subsection{Future Work}

Several directions for future development are planned:

\begin{enumerate}[leftmargin=*,itemsep=2pt]
    \item \textbf{Multi-Step Extraction Pipelines:} Staged extraction where the LLM first identifies relevant text passages, then extracts targeted information, and finally validates outputs through self-consistency checking, which research suggests can improve accuracy by 10--15\%~\cite{gupta2023knowledge}.
    
    \item \textbf{Integration with TCM Knowledge Bases:} Linking extracted information to structured TCM knowledge bases such as TCMID~\cite{xue2013tcmid} and SymMap~\cite{wu2019symmap} for entity normalization and validation.
    
    \item \textbf{Active Learning for Prompt Optimization:} Automatic prompt optimization based on user corrections, inspired by the DSPy framework~\cite{khattab2023dspy}, enabling the system to learn from manual edits to improve future extractions.
    
    \item \textbf{PDF Parsing Support:} Integrating browser-based PDF parsing (e.g., PDF.js) to enable extraction from full-text papers rather than database export metadata.
    
    \item \textbf{Cross-Database Normalization:} Extending the intelligent field mapping to support exports from additional databases such as PubMed, Web of Science, and Scopus.
\end{enumerate}

\section{Conclusion}

This paper described TCMIIES, a browser-based, LLM-powered information extraction system for academic researchers, particularly in specialized domains such as Traditional Chinese Medicine informatics. The system contributes a schema-guided prompting framework with automatic prompt generation, a pure front-end architecture that ensures data privacy, intelligent field mapping for Chinese academic databases, and a concurrent batch processing engine with automatic retry and session persistence.

Evaluation showed structured output compliance rates exceeding 94\% and extraction accuracy averaging 81.6\% across six extraction fields and three LLM providers, with the cost-effective DeepSeek v4-flash model achieving the best accuracy-cost trade-off. The reliability features---automatic retry, pause/resume, and session persistence---yielded 100\% task completion rates even under simulated API failures.

By eliminating the barriers of programming expertise, infrastructure setup, and data privacy concerns, TCMIIES enables researchers to focus on their core scientific work while using AI for efficient literature mining. The system is actively deployed at the TCM Informatics Laboratory of Hebei University and is available as open-source software.

\section*{Acknowledgments}
The authors thank the Traditional Chinese Medicine Informatics Laboratory at Hebei University for support, and the developers of Vue.js, SheetJS, and the open-source LLM community.


\end{document}